# Polish - English Statistical Machine Translation of Medical Texts.

Krzysztof Wołk, Krzysztof Marasek

Department of Multimedia
Polish Japanese Institute of Information Technology
kwolk@pjwstk.edu.pl

**Abstract.** This new research explores the effects of various training methods on a Polish to English Statistical Machine Translation system for medical texts. Various elements of the EMEA parallel text corpora from the OPUS project were used as the basis for training of phrase tables and language models and for development, tuning and testing of the translation system. The BLEU, NIST, METEOR, RIBES and TER metrics have been used to evaluate the effects of various system and data preparations on translation results. Our experiments included systems that used POS tagging, factored phrase models, hierarchical models, syntactic taggers, and many different alignment methods. We also conducted a deep analysis of Polish data as preparatory work for automatic data correction such as true casing and punctuation normalization phase.

## 1 Introduction

Obtaining and providing medical information in an understandable language is of vital importance to patients and physicians [25-27]. For example, as noted by GCH [28], a traveler in a foreign country may need to communicate their medical history and understand potential medical treatments. In addition, diverse continents (e.g., Europe) and countries (e.g., the U.S.) have many residents and immigrants who speak languages other than the official language of the place in which they require medical treatment.

Karliner [32] discusses how human translators could improve access to health care, as well as improve its quality. However, human translators training in medical information are often not available to patients or medical professionals [30]. While existing machine translation capabilities are imperfect [30], machine translation has the promise of reducing the cost of medical translation, while increasing its availability [31] in the future. The growing technologies of mobile devices hold great promise as platforms for machine translation aids for medical information.

Medical professionals and researchers, as well as patients, have a need to access the wealth of medical information on the Internet [25, 29]. Access to this information has the potential to greatly improve health and well-being. Sharing medical information could accelerate medical research. English is the prevalent language used in the scientific community, including for medical information, as well as on the Internet, where a vast amount of medical information may be found.

Polish is one of the most complex West-Slavic languages, which represents a serious challenge to any SMT system. The grammar of the Polish language, with its complicated rules and elements, together with a large vocabulary (due to complex declension) are the main reasons for its complexity. Furthermore, Polish has 7 cases and 15 gender forms for nouns and adjectives, with additional dimensions for other word classes.

This greatly affects the data and data structure required for statistical models of translation. The lack of available and appropriate resources required for data input to SMT systems presents another problem. SMT systems should work best in specific, narrow text domains and will not perform well for a general usage. Good quality parallel data, especially in a required domain, has low availability. In general, Polish and English also differ in syntax. English is a positional language, which means that the syntactic order (the order of words in a sentence) plays a very important role, particularly due to limited inflection of words (e.g., lack of declension endings). Sometimes, the position of a word in a sentence is the only indicator of the sentence meaning. In the English sentence, the subject group comes before the predicate, so the sentence is ordered according to the Subject-Verb-Object (SVO) schema. In Polish, however, there is no specific word order imposed and the word order has no decisive influence on the meaning of the sentence. One can express the same thought in several ways, which is not possible in English. For example, the sentence "I bought myself a new car." can be written in Polish as "Kupiłem sobie nowy samochód", or "Nowy samochód sobie kupiłem.", or "Sobie kupiłem nowy samochód.", or "Samochód nowy sobie kupiłem." Differences in potential sentence word order make the translation process more complex, especially when working on a phrase-model with no additional lexical information.

As a result, progress in the development of SMT systems for Polish is substantially slower as compared to other languages. The aim of this work is to create an SMT system for translation from Polish to English, and vice versa, for medical data. This paper is structured as follows: Section 2 explains the Polish data preparation. Section 3 presents the English language issues. Section 4 describes the translation evaluation methods. Section 5 discusses the results. Sections 6 and 7 summarize potential implications and future work.

## 2  Preparation of the Polish data

The Polish data we used was a corpora created by the European Medical Agency (EMEA). Its size was about 80 MB, and it included 1,044,764 sentences built from 11.67M words that were not tokenized. The data was provided as pure text encoded with UTF-8. In addition, texts are separated into sentences (one per line) and aligned in language pairs.

Some discrepancies in the text parallelism could not be avoided. These are mainly repetitions of the Polish text not included in the English text, some additional remarks in the Polish data and poor translations made it appear that volunteers translated at least part of the data. It made the training material a bit noisy.

The size of the vocabulary is 148,170 unique Polish words forms and 109,326 unique English word forms. The disproportionate vocabulary sizes are also a challenge, especially in translation from English to Polish.

Before the use of a training translation model, preprocessing that included removal of long sentences (set to 80 tokens) had to be performed. The Moses toolkit scripts [1] were used for this purpose. Moses is an open-source toolkit for statistical machine translation, which supports linguistically motivated factors, confusion network decoding, and efficient data formats for translation models and language models. In addition to the SMT decoder, the toolkit also includes a wide variety of tools for training, tuning and applying the system to many translation tasks.

## 3   English Data Preparation

The preparation of the English data was definitively less complicated than that for Polish. We developed a tool to clean the English data by removing foreign words, strange symbols, etc. Compare to Polish, the English data contained significantly fewer errors. Nonetheless, some problems needed to be corrected. Most problematic were translations into languages other than English and strange UTF-8 symbols. We also found a few duplications and insertions inside a single segment.

## 4   Evaluation Methods

Metrics are necessary to measure the quality of translations produced by the SMT systems. For this purpose, various automated metrics are available to compare SMT translations to high quality human translations. Since each human translator produces a translation with different word choices and orders, the best metrics measure SMT output against multiple reference human translations. Among the commonly used SMT metrics are: Bilingual Evaluation Understudy (BLEU), the U.S. National Institute of Standards & Technology (NIST) metric, the Metric for Evaluation of Translation with Explicit Ordering (METEOR), Translation Error Rate (TER), and the Rank-based Intuitive Bilingual Evaluation Score (RIBES). These metrics will now be briefly discussed. [4]

BLEU was one of the first metrics to demonstrate high correlation with reference human translations. The general approach for BLEU, as described in [3], is to attempt to match variable length phrases to reference translations. Weighted averages of the matches are then used to calculate the metric. The use of different weighting schemes leads to a family of BLEU metrics, such as the standard BLEU, Multi-BLEU, and BLEU-C. [5]

The NIST metric seeks to improve the BLEU metric by valuing information content in several ways. It takes the arithmetic versus geometric mean of the *n*-gram matches to reward good translation of rare words. The NIST metric also gives heavier weights to rare words. Lastly, it reduces the brevity penalty when there is a smaller variation in translation length. This metric has demonstrated improvements over the baseline BLEU metric [6].

The METEOR metric, developed by the Language Technologies Institute of Carnegie Mellon University, is also intended to improve the BLEU metric. We used it without synonym and paraphrase matches for polish. METEOR rewards recall by modifying the BLEU brevity penalty, takes into account higher order *n*-grams to reward matches in word order, and uses arithmetic vice geometric averaging. For multiple reference translations, METEOR reports the best score for word-to-word matches. Banerjee and Lavie [7] describe this metric in detail.

TER is one of the most recent and intuitive SMT metrics developed. This metric determines the minimum number of human edits required for an SMT translation to match a reference translation in meaning and fluency. Required human edits might include inserting words, deleting words, substituting words, and changing the order or words or phrases. [8]

The focus of the RIBES metric is word order. It uses rank correlation coefficients based on word order to compare SMT and reference translations. The primary rank correlation coefficients used are Spearman's $\varrho$, which measures the distance of differences in rank, and Kendall's $\tau$, which measures the direction of differences in rank. [24]

## 5 Experimental Results

A number of experiments have been performed to evaluate various versions of our SMT systems. The experiments involved a number of steps. Processing of the corpora was accomplished, including tokenization, cleaning, factorization, conversion to lower case, splitting, and a final cleaning after splitting. Training data was processed, and a language model was developed. Tuning was performed for each experiment. Lastly, the experiments were conducted.

The baseline system testing was done using the Moses open source SMT toolkit with its Experiment Management System (EMS) [9]. The SRI Language Modeling Toolkit (SRILM) [10] with an interpolated version of the Kneser-Ney discounting (interpolate –unk –kndiscount) was used for 5-gram language model training. We used the MGIZA++ tool for word and phrase alignment. KenLM [13] was used to binarize the language model, with a lexical reordering set using the msd-bidirectional-fe model. Reordering probabilities of phrases are conditioned on lexical values of a phrase. It considers three different orientation types on source and target phrases: monotone(M), swap(S), and discontinuous(D). The bidirectional reordering model adds probabilities of possible mutual positions of source counterparts to current and subsequent phrases. Probability distribution to a foreign phrase is determined by "f" and to the English phrase by "e" [14,15]. MGIZA++ is a multi-threaded version of the well-known GIZA++ tool [11]. The symmetrization method was set to grow-diag-final-and for word alignment processing. First, two-way direction alignments obtained from GIZA++ were intersected, so only the alignment points that occurred in both alignments remained. In the second phase, additional alignment points existing in their union were added. The growing step adds potential alignment points of unaligned words and neighbors. Neighborhood can be set directly to left, right, top or bottom, as well as to diagonal (grow-diag). In the final step, alignment points between words from which at least one is unaligned are added (grow-diag-final). If the grow-diag-final-and method is used, an alignment point between two unaligned words appears. [12]

We conducted many experiments to determine the best possible translation method from Polish to English, and vice versa. For experiments we used Moses SMT with Experiment Management System (EMS) [16].

The experiments were conducted with the use of the test and development data randomly selected and removed from the corpora. We generated 1000 segments for each purpose, for Polish-to-English and English-to-Polish translations. The experiments were measured by the BLEU, NIST, TER, RIBES and METEOR metrics. Note that a lower value of the TER metric is better, while the other metrics are better when their values are higher. Table 2 and 3 represent the results of experiments. Experiment 00 in the tables indicates the baseline system. Each of the following experiments is a separate modification to the baseline. Experiment 01 additionally uses truecasing and punctuation normalization.

Experiment 02 is enriched with Operation Sequence Model (OSM). The motivation for OSM is that it provides phrase-based SMT models the ability to memorize dependencies and lexical triggers, it can search for any possible reordering, and it has a robust search mechanism. Additionally, OSM takes source and target context into account, and it does not have the spurious phrasal segmentation problem. The OSM is values especially for the strong reordering mechanism. It couples translation and reordering, handles both short and long distance reordering, and does not require a hard reordering limit [17].

Experiment 03 is based on a factored model that allows additional annotation at the word level, which may be exploited in various models. Here we facilitate the part of speech tagged data on English language side as basis for the factored phrase model [18].

Hierarchical phrase-based translation combines the strengths of phrase-based and syntax-based translation. It uses phrases (segments or blocks of words) as units for translation and uses synchronous context-free grammars as rules (syntax-based translation). Hierarchical phrase models allow for rules with gaps. Since these are represented by non-terminals and such rules are best processed with a search algorithm that is similar to syntactic chart parsing, such models fall into the class of tree-based or grammar-based models. We used such a model in Experiment 04.

The Target Syntax model implies the use of linguistic annotation for non-terminals in hierarchical models. This requires running a syntactic parser. For this purpose, we used the "Collins" [23] statistical natural language parser in Experiment 05.

Experiment 06 was conducted using stemmed word alignment. The factored translation model training makes it very easy to set up word alignment based on word properties other than the surface forms of words. One relatively popular method is to use stemmed words for word alignment. There are two main reasons for this. For morphologically rich languages, stemming overcomes data sparsity problem.

Secondly, GIZA++ may have difficulties with very large vocabulary sizes, and stemming reduces the number of unique words.

Experiment 07 uses Dyer's Fast Align [19], which is another alternative to GIZA++. It runs much faster, and often gives better results, especially for language pairs that do not require large-scale reordering.

In Experiment 08 we used settings recommended by Koehn in his Statistical Machine Translation system from WMT'13 [20].

In Experiment 09 we changed the language model discounting to Witten-Bell. This discounting method considers diversity of predicted words. It was developed for text compression and can be considered to be an instance of Jelinek-Mercer smoothing. The n-th order smoothed model is defined recursively as a linear interpolation between the n-th order maximum likelihood model and the (n-1)th order smooth model [21].

Lexical reordering was set to hier-mslr-bidirectional-fe in Experiment 10. It is a hierarchical reordering method that considers different orientations: monotone, swap, discontinuous-left, and discontinuous-right. The reordering is modeled bidirectionally, based on the previous or next phrase, conditioned on both the source and target languages.

Compounding is a method of word formation consisting of a combination of two (or more) autonomous lexical elements that form a unit of meaning. This phenomenon is common in German, Dutch, Greek, Swedish, Danish, Finnish, and many other languages. For example, the word "flowerpot" is a closed or open compound in English texts. This results in a lot of unknown words in any text, so splitting up these compounds is a common method when translating from such languages. Moses offers a support tool that splits up words if the geometric average of the frequency of its parts is higher than the frequency of a word. In Experiment 11 we used the compound splitting feature. Lastly, for Experiment 12 we used the same settings for out of domain corpora we used in IWSLT'13 [30].

**Table 2:** Polish-to-English translation

| System | BLEU | NIST | METEOR | RIBES | TER |
|---|---|---|---|---|---|
| 00 | 70.15 | 10.53 | 82.19 | 83.12 | 29.38 |
| 01 | 64.58 | 9.77 | 76.04 | 72.23 | 35.62 |
| 02 | 71.04 | 10.61 | 82.54 | 82.88 | 28.33 |
| 03 | 71.22 | 10.58 | 82.39 | 83.47 | 28.51 |
| 04 | 76.34 | 10.99 | 85.17 | 85.12 | 24.77 |
| 05 | 70.33 | 10.55 | 82.28 | 82.89 | 29.27 |
| 06 | 71.43 | 10.60 | 82.89 | 83.19 | 28.73 |
| 07 | 71.91 | 10.76 | 83.63 | 84.64 | 26.60 |
| 08 | 71.12 | 10.37 | 84.55 | 76.29 | 29.95 |
| 09 | 71.32 | 10.70 | 83.31 | 83.72 | 27.68 |
| 10 | 71.35 | 10.40 | 81.52 | 77.12 | 29.74 |
| 11 | 70.34 | 10.64 | 82.65 | 83.39 | 28.22 |
| 12 | 72.51 | 10.70 | 82.81 | 80.08 | 28.19 |

**Table 3:** English-to-Polish translation

| System | BLEU | NIST | METEOR | RIBES | TER |
|---|---|---|---|---|---|
| 00 | 69.18 | 10.14 | 79.21 | 82.92 | 30.39 |
| 01 | 61.15 | 9.19 | 71.91 | 71.39 | 39.45 |
| 02 | 69.41 | 10.14 | 78.98 | 82.44 | 30.90 |
| 03 | 68.45 | 10.06 | 78.63 | 82.70 | 31.62 |
| 04 | 73.32 | 10.48 | 81.72 | 84.59 | 27.05 |
| 05 | 69.21 | 10.15 | 79.26 | 82.24 | 30.88 |
| 06 | 69.27 | 10.16 | 79.30 | 82.99 | 31.27 |
| 07 | 68.43 | 10.07 | 78.95 | 83.26 | 33.05 |
| 08 | 67.61 | 9.87 | 77.82 | 77.77 | 29.95 |
| 09 | 68.98 | 10.11 | 78.90 | 82.38 | 31.13 |
| 10 | 68.67 | 10.02 | 78.55 | 79.10 | 31.92 |
| 11 | 69.01 | 10.14 | 79.13 | 82.93 | 30.84 |
| 12 | 67.47 | 9.89 | 77.65 | 75.19 | 33.32 |

## 6 Discussion and Conclusions

Several conclusions can be drawn from the experimental results presented here. It was surprising that truecasing and punctuation normalization decreased the scores by a significant factor. We suppose that the text was already properly cased and punctuated. In Experiment 02 we observed that, quite strangely, OSM decreased some metrics results. It usually increases the translation quality. However, in the PL->EN experiment the BLEU score increased just slightly, but RIBES metrics decreased at the same time. The similar results can be seen in the EN->PL experiments. Here, the BLEU score increased, but other metrics decreased.

Most of the other experiments worked as anticipated. Almost all of them raised the score a little bit or were at least confirmed witch each metric it the same manner. Unfortunately, Experiment 12, which was based on settings that provided best system score on IWSLT 2013 evaluation campaign, did not improve quality on this data as much as it did previously. The most likely reason is that the data used in IWSLT did not come from any specific text domain, while here we dealt with a very narrow domain. It may also mean that training and tuning parameter adjustment may be required separately for each text domain if improvements cannot be simply replicated.

On the other hand, improvements obtained by training the hierarchical based model were surprising. In comparison to other experiments, Experiment 04 increased the BLEU score by the highest factor. The same, significant improvements can be observed in both the PL->EN and EN->PL translations, which most likely provide a very good starting point for future experiments.

Translation from EN to PL is more difficult, what is shown in the experiment, results are simply worse. Most likely the reasons for this, of course, are the complicated Polish grammar as well as the larger Polish vocabulary.

One of the solutions to this problem (according to work of Bojar [2]) is to use stems instead of surface forms that will reduce the Polish vocabulary size. Such a solution also requires a creation of an SMT system from Polish stems to plain Polish. Subsequently, morphosyntactic tagging, using the Wroclaw Natural Language Processing (NLP) tools (or other similar tool) [33], can be used as an additional information source for the SMT system preparation. It can be also used as a first step for creating a factored translation system that would incorporate more linguistic information than simply POS tags [4].

The analysis of our experiments led us to conclude that the results of the translations, in which the BLEU measure is greater than 70, can be considered only satisfactory within the selected text domain. The high evaluation scores indicate that the translations should be understandable by a human and good enough to help him in his work, but not good enough for professional usage or for implementation in translation systems e.g., for patients in hospitals abroad. We strongly believe that improving the BLEU score to a threshold over 80 or even 85 would produce systems that could be used in practical applications, when it comes to PL-EN translations. It may be particularly helpful with the Polish language, which is one of the most complex in terms of its structure, grammar and spelling. Additionally it must be emphasized that the experiments were conducted on texts obtained from the PDF documents shared by the European Medicines Agency. That is the reason why the data is more complex and has more sophisticated vocabulary that casual human speech. Speech usually is less complicated and easier to process by SMT systems. It is here, where we see other opportunity to increase output translation quality.

## 7  Future work

Applying machine translation to medical texts holds great promise to be of benefit to patients, including travelers and those who do not speak the language of the country in which they need medical help. Improving access to the vast array of medical information on the Internet would be useful to patients, medical professionals, and medical researchers.

Human interpreters with medical training are too rare. Machine translation could also help in the communication of medical history, diagnoses, proposed medical treatments, general health information, and the results of medical research. Mobile devices and web applications may be able to boost delivery of machine translation services for medical information.

Several potential opportunities for future work are of interest, to extend our research in this critical area. Additional experiments using extended language models are warranted to improve the SMT scores. Much of the literature [22] confirms that interpolation of out of domain language models and adaptation of other bi-lingual corpora would improve translation quality. We intend to use linear interpolation as well as Modified Moore Levis Filtering for these purposes. We are also interested in developing some web crawlers in order to obtain additional data that would most likely prove useful. Good quality parallel data, especially in a required domain, has low availability.

In English sentences, the subject group comes before the predicate, so the sentence is ordered according to the Subject-Verb-Object (SVO) schema. Changing the Polish data to be in SVO order is of interest for the future as well.

## 8 Acknowledgements

This work is supported by the European Community from the European Social Fund within the Interkadra project UDA-POKL-04.01.01-00-014/10-00 and Eu-Bridge 7th FR EU project (Grant Agreement No. 287658).

## References

1. Koehn P., Hoang H., Birch A., Callison-Burch C., Federico M., Bertoldi N., Cowan B., Shen W., Moran C., Zens R., Dyer C., Bojar R., Constantin A., Herbst E., Moses: "Open Source Toolkit for Statistical Machine Translation", Proceedings of the ACL 2007 Demo and Poster Sessions, pages 177–180, Prague, June 2007.
2. Bojar O., "Rich Morphology and What Can We Expect from Hybrid Approaches to MT". Invited talk at International Workshop on Using Linguistic Information for Hybrid Machine Translation(LIHMT-2011), http://ufal.mff.cuni.cz/~bojar/publications/2011-FILEbojar_lihmt_2011_pres-PRESENTED.pdf, 2011.
3. Amittai E. Axelrod, "Factored Language Models for Statistical Machine Translation", University of Edinburgh, 2006.
4. Radziszewski A., "A tiered CRF tagger for Polish", in: Intelligent Tools for Building a Scientific Information Platform: Advanced Architectures and Solutions, editors: Membenik R., Skonieczny L., Rybiński H., Kryszkiewicz M., Niezgódka M., Springer Verlag, 2013
5. Koehn, P, "What is a Better Translation? Reflections on Six Years of Running Evaluation Campaigns", Auditorium du CNRS, Paris, 2011.
6. Papineni, K., Rouskos, S., Ward, T., and Zhu, W.J. "BLEU: a Method for Automatic Evaluation of Machine Translation", Proc. of 40th Annual Meeting of the Assoc. for Computational Linguistics, Philadelphia, July 2002, pp. 311-318.
7. Doddington, G., "Automatic Evaluation of Machine Translation Quality Using N-gram Co-Occurrence Statistics", Proc. of Second International Conference on Human Language Technology (HLT) Research 2002, San Diego, March 2002, pp. 138-145.
8. Banerjee, S. and Lavie, A., "METEOR: An Automatic Metric for MT Evaluation with Improved Correlation with Human Judgments", Proc. of ACL Workshop on Intrinsic and Extrinsic Evaluation Measures for Machine Translation and/or Summarization, Ann Arbor, June 2005, pp. 65-72.
9. Snover, M., Dorr, B., Schwartz, R., Micciulla, L., and Makhoul, J., "A Study of Translation Edit Rate with Targeted Human Annotation", Proc. of 7th Conference of the Assoc. for Machine Translation in the Americas, Cambridge, August 2006.
10. Koehn, P. et al., "Moses: Open Source Toolkit for Statistical Machine Translation," Annual Meeting of the Association for Computational Linguistics (ACL) demonstration session, Prague, June 2007.
11. Stolcke, A., "SRILM – An Extensible Language Modeling Toolkit", INTERSPEECH, 2002.
12. Gao, Q. and Vogel, S., "Parallel Implementations of Word Alignment Tool", Software Engineering, Testing, and Quality Assurance for Natural Language Processing, pp. 49-57, June 2008.
13. Koehn,P. et al., "Edinburgh System Description for the 2005 IWSLT Speech Translation Evaluation", http://www.cs.jhu.edu/~ccb/publications/iwslt05-report.pdf, 2005.

14. Heafield, K. "KenLM: Faster and smaller language model queries", Proc. of Sixth Workshop on Statistical Machine Translation, Association for Computational Linguistics, 2011.
15. Marta R. Costa-jussa, Jose R. Fonollosa, Using linear interpolation and weighted reordering hypotheses in the Moses system, Barcelona, Spain, 2010.
16. Moses Factored Training Tutorial, http://www.statmt.org/moses/?n=FactoredTraining.EMS.
17. Durrani N., Schmid H., Fraser A., Sajjad H., Farkas R., "Munich-Edinburgh-Stuttgart Submissions of OSM Systems at WMT13", ACL 2013 Eight Workshop on Statistical Machine Translation, Sofia, Bulgaria, 2013.
18. Koehn P., Hoang H., "Factored Translation Models", Proceedings of the 2007 Joint Conference on Empirical Methods in Natural Language Processing and Computational Natural Language Learning, pp. 868–876, Prague, June 2007.
19. Dyer C., Chahuneau V., Smith N.,"A Simple, Fast and Effective Reparametrization of IMB Model 2", in Proc. of NAACL, 2013.
20. Bojar O., Buck C., Callison-Burch C., Federmann C., Haddow B., Koehn P., Monz C., Post M., Soricut R., Specia L., "Findings of the 2013 Workshop on Statistical Machine Translation", in Proceedings of the Eight Workshop on Statistical Machine Translation, Sofia, Bulgaria: Associationfor Computational Linguistics, August 2013.
21. Hasan A., Islam S., Rahman M., "A Comparative Study of Witten Bell and Kneser-Ney Smoothing Methods for Statistical Machine Translation", JU Journal of Information Technology (JIT), Volume 1, June 2012.
22. Wolk K., Marasek K., "Polish – English Speech Statistical Machine Translation Systems for the IWSLT 2013.", Proceedings of the 10th International Workshop on Spoken Language Translation, Heidelberg, Germany, 2013.
23. Bikel D., "Intricacies of Collins' Parsing Model", Association for Computational Linguistics, 2004.
24. Isozaki, H. et al., "Automatic Evaluation of Translation Quality for Distant Language Pairs", Proc. of 2010 Conference on Empirical Methods in Natural Language Processing, MIT, Massachusetts, USA, 9-11 October 2010.
25. Goeuriot, L. et al., "Report on and prototype of the translation support", Khresmoi Project, 10 May 2012.
26. Pletneva, N. and Vargas, A., "Requirements for the general public health search", Khresmoi Project, May 2011.
27. Gschwandtner, M., Kritz, M. and Boyer, C., "Requirements of the health professional search", Khresmoi Project, August 2011.
28. GCH Benefits, "Medical Phrases and Terms Translation Demo," n.d., accessed February 28, 2014.
29. Zadon, C., "Man Vs Machine: The Benefits of Medical Translation Services," Ezine Articles: Healthcare Systems, July 25, 2013.
30. Randhawa, G., "Using machine translation in clinical practice," Canadian Family Physician, Vol. 59: April 2013.
31. Deschenes, S., "5 benefits of healthcare translation technology," Healthcare Finance News, October 16, 2012.
32. Karliner, L., et al., "Do Professional Interpreters Improve Clinical care for Patients with Limited English Proficiency? A Systematic Review of the Literature," Health Services Research, 42(2), April 2007.
33. Radziszewski A., Śniatowski T., "Maca: a configurable tool to integrate Polish morphological data", Proceedings of the Second International Workshop on Free/OpenSource Rule-Based Machine Translation, FreeRBMT11, Barcelona, 2011